\documentclass{article}
\usepackage{spconf,amsmath,graphicx}
\usepackage{amssymb}
\usepackage{xcolor}
\usepackage{icomma}
\usepackage{bm}
\usepackage{multirow}
\def\SSSD{{\sc sssd}}

\usepackage{hyperref}
\title{Smooth and Stepwise Self-Distillation for Object Detection}
\name{Jieren Deng$^{\dag}$\textsuperscript{\textsection}\thanks{The work was done when the author was an intern at Baidu Research.}, Xin Zhou$^{\dag}$, Hao Tian$^{\dag}$, Zhihong Pan$^{\dag}$, Derek Aguiar\textsuperscript{\textsection}}

\address{\textsuperscript{\textsection}University of Connecticut, CT, USA \\ $^{\dag}$Baidu Research USA, Sunnyvale, CA, USA}

\begin{document}
%\ninept
%
\maketitle
\begin{abstract}
% The abstract should appear at the top of the left-hand column of text, about
% 0.5 inch (12 mm) below the title area and no more than 3.125 inches (80 mm) in
% length.  Leave a 0.5 inch (12 mm) space between the end of the abstract and the
% beginning of the main text.  The abstract should contain about 100 to 150
% words, and should be identical to the abstract text submitted electronically
% along with the paper cover sheet.  All manuscripts must be in English, printed
% in black ink.

Distilling the structured information captured in feature maps has contributed to improved results for object detection tasks, but requires careful selection of baseline architectures and substantial pre-training.
Self-distillation addresses these limitations and has recently achieved state-of-the-art performance for object detection despite making several simplifying architectural assumptions.
Building on this work, we propose \textbf{S}mooth and \textbf{S}tepwise \textbf{S}elf-\textbf{D}istillation (\SSSD{}) for object detection.
Our \SSSD{} architecture forms an implicit teacher from object labels and a feature pyramid network backbone to distill label-annotated feature maps using Jensen-Shannon distance, which is smoother than distillation losses used in prior work.
We additionally add a distillation coefficient that is adaptively configured based on the learning rate.
We extensively benchmark \SSSD{} against a baseline and two state-of-the-art object detector architectures on the COCO dataset by varying the coefficients and backbone and detector networks.
We demonstrate that \SSSD{} achieves higher average precision in most experimental settings, is robust to a wide range of coefficients, and benefits from our stepwise distillation procedure.

%Recently, self-distilled object detection models have achieved state-of-the-art performance,  

% They use MSE for the feature maps in self distillation
% Prior methods use \lambda=1
% when learning rate decreases, we increase \lambda which weights KD

\end{abstract}
\section{Introduction}
\label{sec:intro}

% These guidelines include complete descriptions of the fonts, spacing, and
% related information for producing your proceedings manuscripts. Please follow
% them and if you have any questions, direct them to Conference Management
% Services, Inc.: Phone +1-979-846-6800 or email
% to \\\texttt{papers@2021.ieeeicassp.org}.

% DEREK TODO:
Knowledge distillation is a technique for transferring the information contained in the feature maps and model outputs of a large \textit{teacher} model to a typically smaller \textit{student} model~\cite{KD_hinton,bucilua2006model}.
As a result, student models have lower storage and memory requirements and yield more efficient inference, enabling use in limited resource or real-time settings like in edge devices or autonomous vehicles~\cite{bharadhwaj2022detecting,kothandaraman2021domain}.
Object detection is among the largest beneficiary of knowledge distillation~\cite{Li_Li_Wang_Zhang_Wu_Liang_2022,NEURIPS2021_892c91e0,zhang2021improve} and transfer learning on related tasks~\cite{freezeFeature}, but these techniques require careful selection of a baseline teacher model and expensive pre-training~\cite{zhang2021improve,yao2021g}. 
Recent work removes the dependency on a pre-trained teacher entirely, e.g. by collaboratively training a collection of student networks (collaborative learning)~\cite{guo2020online} or smoothing class labels (label regularization)~\cite{muller2019does,ding2019adaptive}; however, these methods have largely focused on image classification.

% \noindent \textbf{Knowledge Distillation}
% ...... However, in real-world, a valid and pre-trained teacher is not always available and accessible. \\

% \noindent \textbf{Self-distillation}
% While freezing the layers of a pre-trained backbone on a related task (e.g. image classification) has been shown to improve object detection in a transfer learning setting~\cite{freezeFeature},  demonstrates that a pre-trained model 
% \textbf{While freezing the layers of a pre-trained backbone model to preserve the features learned on a related task (e.g., image classification) improves object detection~\cite{freezeFeature}.
% However, recent researches~\cite{Li_Li_Wang_Zhang_Wu_Liang_2022,NEURIPS2021_892c91e0,zhang2021improve,Zhang2022LGDLS,labelEnc} on knowledge distillation show that the model can be improved by minimizing the difference of extracted feature from backbone between other knowledge. In another word, enhancing the extracting performance of backbone instead of freezing backbone can effectively improve the performance of the model.}
Unlike traditional transfer learning and knowledge distillation, \textit{self-distillation} aims at extracting knowledge from the data labels during feature extraction within the same backbone model~\cite{Li_Li_Wang_Zhang_Wu_Liang_2022,NEURIPS2021_892c91e0,zhang2021improve,Zhang2022LGDLS,labelEnc}; this eliminates the need for expensive pre-training of a teacher network.
LabelEnc is a recently developed self-distillation method for object detection that encodes label information within the feature maps, providing intermediate supervision at internal neural network layers and achieving an approximately 2\% improvement over prior work in the COCO dataset~\cite{labelEnc}.
Building on LabelEnc, label-guided self-distillation (LGD) leverages both label- and feature map-encodings as knowledge and improved the benchmark set by LabelEnc on COCO~\cite{Zhang2022LGDLS}.

\begin{figure*}[h]
\centering
\includegraphics[width=0.85\textwidth]{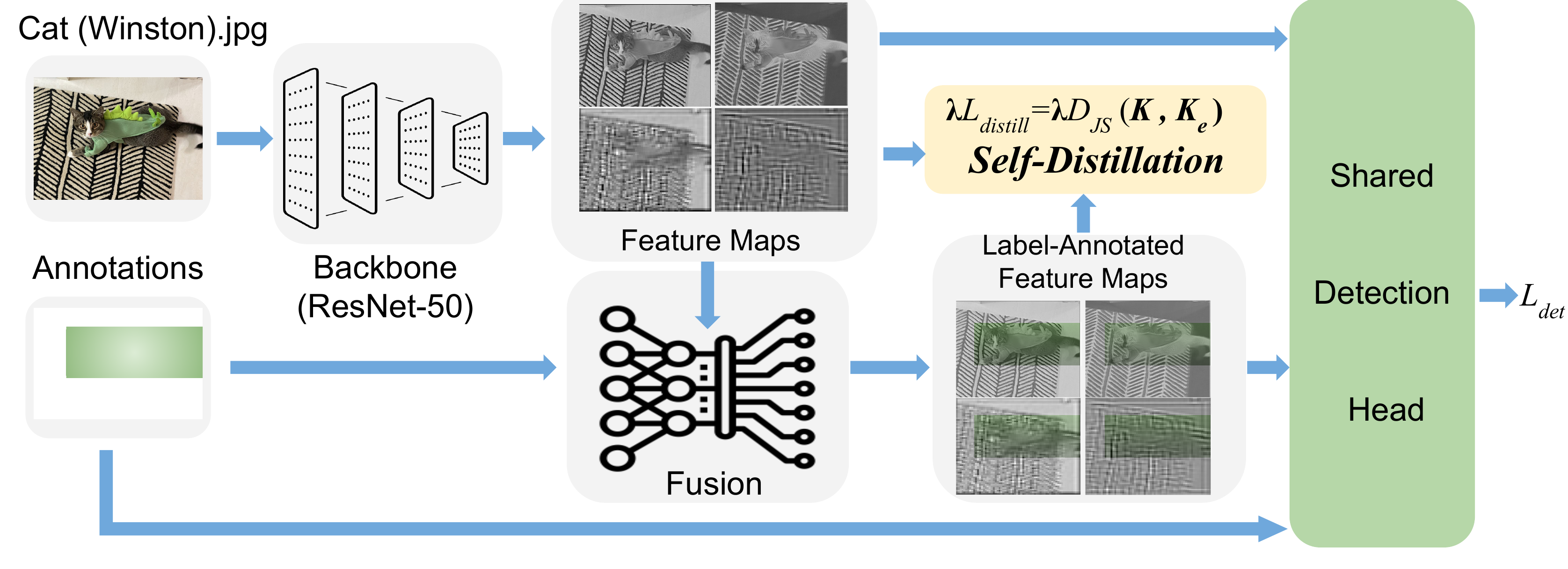}
\caption{\textbf{\textbf{S}mooth and \textbf{S}tepwise \textbf{S}elf-\textbf{D}istillation (\SSSD{}).} The feature maps ($\bm{K}$) extracted from the backbone (ResNet-50) are sent to the fusion component along with the ground truth annotations. 
The distillation loss ($L_{distill}$) is calculated using the feature maps and label enhanced feature maps ($\bm{K_e}$). 
The detection loss ($L_{det}$) is calculated as classification and bounding-box regression losses by a shared detection head.}
\label{Fig:overall} 
\end{figure*}

While LabelEnc and LGD achieve state-of-the-art performance, they make simplifying
architectural assumptions.
First, they consider mean squared error (MSE) as the only distillation loss, which is not robust to the noisy or imperfect teachers that are commonplace in self-distillation settings~\cite{kim2021comparing}. %TODO: add text and citation for the limitations of MSE in general or specifically in KD
Second, there is no consideration for how the knowledge distillation coefficient $\lambda$ affects the total loss or overall performance. %TODO: add test and citations for the importance of carefully selection \lambda
In this paper, we explore the limitations of MSE as a self-distillation loss and the sensitivity of self-distillation to $\lambda$.
%and the unexploited knowledge from the data and improving the exploitation of the data to excavate knowledge that is beneficial to the model without seeking the help from a teacher model. 
We propose \textbf{S}mooth and \textbf{S}tepwise \textbf{S}elf-\textbf{D}istillation (\SSSD{}) by combining the Jensen-Shannon (JS) divergence with a $\lambda$ that is adaptively configured based on the learning rate in a stepwise manner~(Fig.~\ref{Fig:overall}).
We summarize our contributions as follows:

% Computer vision tasks like image classification~\cite{xu2020knowledge, 10.5555/3495724.3496961, chen2021cross} and object detection are among the largest beneficiaries of knowledge distillation, 

\begin{itemize}
    \item We present \textbf{S}mooth and \textbf{S}tepwise \textbf{S}elf-\textbf{D}istillation (\SSSD{}), which combines  stepwise self-distillation with a smooth, bounded, and symmetric distance that is robust to noise (JS)~\cite{xu2019l_dmi,wei2020optimizing,englesson2021generalized}.
    \item  We study the sensitivity of self-distillation to the distillation coefficient $\lambda$ under a variety of architectural assumptions, providing insight on how $\lambda$ influences model performance.
    \item We thoroughly benchmark \SSSD{} and demonstrate higher average precision than previous self-distillation approaches in most configurations of the backbone and detector networks.
\end{itemize}

% \section{Background}
% \subsection{Knowledge Distillation}
% \subsection{Knowledge Distillation without a Teacher}
% \label{sec:background}

\section{Proposed Method}
\label{sec:method}

\subsection{Smooth Self-Distillation}
Leveraging prior work on self-distillation for object detection, the features are obtained from a backbone feature pyramid network with $P$ scales~\cite{labelEnc,Zhang2022LGDLS}.
We define $\bm{K} = \{\bm{k}^{p}\in\mathbb{R}^{N_p \times M_p}\}^{P}$ to be the set of features from the backbone feature pyramid network where $\bm{k}^{p}$ is a vector of features at the $p^{th}$ scale, each pyramid has dimension $N_p \times M_p$, and $p \in \{1,\dots,P\}$. 
Similarly, let $\bm{K}_e = \{\bm{k}^{p}_e\in\mathbb{R}^{N_p \times M_p}\}^{P}$ be the feature maps obtained from a spatial transformer network~\cite{NIPS2015_33ceb07b} (STN) by the label-annotated feature maps (denoted by $e$) in the fusion component (Fig.~\ref{Fig:overall}). 
Existing self-distillation methods for object detection use mean squared error (MSE) to calculate the distillation loss~\cite{Zhang2022LGDLS,labelEnc}: 
\begin{equation*}
    L^{MSE}_{distill} = \frac{1}{N} \sum^{P}_{p=1} || \bm{k}^p - \bm{k}^{p}_e ||^2, 
\end{equation*}
where $N = \sum^{P}_{p=1} N_p \times M_p$ is the total number of feature map elements. 
The Kullback-Leibler (KL) divergence is another commonly used loss function that was used to initially define knowledge distillation~\cite{KD_hinton} and used subsequently across many applications~\cite{8578552,Yang_Xie_Qiao_Yuille_2019,NEURIPS2020_3f5ee243,wang-etal-2021-minilmv2}:
%Therefore, we firstly study if KL divergence could be an alternative of MSE in feature-based knowledge distillation for object detection:
\begin{equation*}
    L^{KL}_{distill} = \frac{1}{N} \sum^{P}_{p=1} D_{KL}(\bm{k}^p || \bm{k}^{p}_e), 
\end{equation*}
However, the KL divergence has several limitations. 
For probability distributions $O$ and $Q$, $D_{KL}(O||Q)$ is not bounded, which may result in model divergence during training, and is sensitive to regions of $O$ and $Q$ that have low probability; e.g., $D_{KL}(O||Q)$ can be large when $O(x) >> Q(x)$ for an event $x$ even if $O(x)$ is small when $Q(x)$ is close to $0$~\cite{berman2020tight}. 
To address these issues, we use the Jensen-Shannon (JS) divergence as a new measure for knowledge distillation in object detection tasks. 
Unlike KL divergence, the JS divergence is bounded by $[0,1]$, symmetric, does not require absolute continuity~\cite{lin1991divergence}, and has been shown to be robust to label noise~\cite{xu2019l_dmi,wei2020optimizing,englesson2021generalized} and imperfect teachers that are commonplace in self-distillation settings~\cite{kim2021comparing}:
%It is a symmetric and smooth version of the KL divergence defined as:
\begin{align*}
    D_{JS}(O||Q) &= D_{JS}(Q||O) \\
    &= \frac{1}{2} D_{KL}(O||M) + \frac{1}{2} D_{KL}(Q||M)
\end{align*}
where $M = \frac{1}{2} (O+Q)$. 
In this work, we consider the JS distance, which is a metric defined by $(D_{JS}(O||Q))^{\frac{1}{2}}$. 
%Research~\cite{js_vs_kl} indicates that JS distance is more generalised and can be understood as being somewhere between $D_{KL}(O || Q) $ and $D_{KL}(Q || O) $. 
% Unlike KL divergence, the JS divergence is bounded:
% \begin{equation}
%    0 \leq D_{JS}(O||Q) \leq 1
% \end{equation}
We define the distillation loss, $L_{distill}$, as: 
\begin{equation*}
    L_{distill} = \frac{1}{N} \sum^{P}_{p=1} \left( D_{JS}(\bm{k}^p || \bm{k}^{p}_e) \right)^{\frac{1}{2}}, 
\end{equation*}
The detection loss is defined as:
\begin{equation*}
    L_{det} = \hat{L}_{det}(H(\bm{K}),Y)+\hat{L}^e_{det}(H(\bm{K}_e),Y)
\end{equation*}
where the $H(\cdot)$ refers to the shared detection head, $Y$ is the ground truth, and $\hat{L}$ is a classification and regression object detection loss. 
Thereby, we obtain the total training objective as:
\begin{equation*}
    L_{total} = L_{det} + \lambda L_{distill}
\end{equation*}
where $\lambda$ is a coefficient for the distillation loss. % i think it's distillation loss here? you can re-delete "loss" if you want
Our choice of functional form for $L_{distill}$ was motivated by research suggesting smooth loss functions improve deep neural network training and performance~\cite{berrada2018smooth,Lapin2016LossFF}; since the JS distance $D_{JS} (0||Q)$ is considered to be a smooth compromise between $D_{KL}(O || Q) $ and $D_{KL}(Q || O) $, we term this knowledge distillation method as \textbf{smooth self-distillation}.
%Research~\cite{berrada2018smooth} shows that smooth  loss function helps the training of deep neural networks and research~\cite{Lapin2016LossFF} also indicates smooth objective is astonishingly competitive. 
%As part of the objective, $L_{distill}$, is also expected be as smooth as possible. 

\subsection{Stepwise Self-Distillation}
Learning rate scheduling is broadly used in large scale deep learning as an important mechanism to adjust the learning rate during training, typically through learning rate reduction according to a predefined schedule. 
%The step of decaying the learning rate is set from 120$k$ iterations to 170$k$ iterations. 
To help the model continue learning from self-distillation during learning rate decay, we propose \textbf{stepwise self-distillation} to compensate for the lessened impact of the self-distillation loss caused by a reduced learning rate. 
In our setting, the backbone model is frozen and the detector is trained in the first 20$k$ iterations. 
An initial $\lambda$ is assigned to the distillation loss empirically after the first 20$k$ iterations; selection of an empirical $\lambda$ is elaborated in the experimental section. 
We redefine the $\lambda$ in stepwise self-distillation as a step function of $\lambda_1$ and a $\lambda_2$ that depends on the training iteration.
Since in our model training the learning rate begins decaying at iteration $120,000$, we define $\lambda$ as:
\begin{equation*}
    \lambda= \begin{cases}
    \lambda_{1},\quad & steps < 120000 \\
    \lambda_{2},\quad & steps \geq 120000
    \end{cases} 
\end{equation*}

\section{Experiments}
We compared \SSSD{} with two state-of-the-art (SOTA) self-distillation architectures for object detection, LabelEnc~\cite{labelEnc} and LGD~\cite{Zhang2022LGDLS}, and a non-distillation baseline model.
All experiments were conducted using the official code repositories for LabelEnc~\cite{labelEncCode} and LGD~\cite{Zhang2022LGDLSCode}, using a batch size of $16$ on $8$ NVIDIA v100 GPUs and configurations specified in their official GitHub repositories.
%Experiments are run with batch size 16 on 8 NVIDIA V100 GPUs. 
%To fairly compare with LGD, we follow the same configuration as released from their Github repository. 
Our experiments tested different backbone networks, ResNet-50 (R-50) and ResNet-101 (R-101), and explored three popular detectors: Faster R-CNN (FRCN)~\cite{NIPS2015_14bfa6bb}, fully convolutional one-stage object detector (FCOS)~\cite{tian2019fcos} and RetinaNet~\cite{retinaNet}. 
All experiments were validated on the Microsoft Common Objects in Context (COCO) dataset with $80$ categories using commonly reported metrics based on mean average precision (AP) and other detailed metrics: APs, APm, and APl, which are the AP for small, medium and large objects, and AP50 and AP75, which are the AP at IoU=0.50 and IoU=0.75 where IoU is the intersection over union~\cite{COCO}.

\subsection{Comparisons with SOTA Results}
We first compared \SSSD{} with competing methods on the COCO data based on AP and using two backbone networks, R-50 and R-101, and three detectors, FRCN, RetinaNet and FCOS (Table~\ref{tab:sota_table}).
Compared to the baseline model, our approach achieved an AP improvement of approximately $2.0\%$, $3.6\%$, and $3.2\%$ for the FRCN, RetinaNet, and FCOS detectors respectively.
%mean AP (mAP) gain for RetinaNet, 3.4\% mAP gain for FCOS and achieves notable advantages in other settings as well. 
Our method improved on the AP of LabelEnc by approximately 2.8\% for FRCN$_{R50}$, 2.2\% for FRCN$_{R101}$, and more than 1\% for other architectural configurations. 
With respect to LGD, \SSSD{} achieves an almost $1\%$ gain in AP for the FRCN$_{R50}$ and FCOS$_{R101}$ configurations and $<0.5\%$ improvements in other FRCN and FCOS settings. 

Since the performance of LGD is most comparable to \SSSD{}, we further investigated the performance of LGD and \SSSD{} using variations of AP (Table~\ref{tab:lgd_table}). 
In the RetinaNet$_{R101}$ setting, our proposed method achieved a $5\%$ AP performance gain ($26.1$ versus $24.9$) for objects with small bounding boxes (APs). 
%The results for the FRCN and FCOS detectors show that \SSSD{} largely outperforms LGD in the AP related metrics.
The results for the other detectors demonstrate that \SSSD{} performs relatively well compared with LGD primarily due to improved AP for objects with medium or large bounding boxes (APm and APl).
FCOS$_{R101}$-based architectures yielded the best AP results for both methods where \SSSD{} outperformed LGD in all AP-related measures besides APs, including a $0.6\%$, $0.8\%$, and $1\%$ gain over LGD in AP50, APm, and APl respectively. % DOUBLE CHECK NUMBERS

% As shown in Table~\ref{tab:sota_table}, we compare our proposed method with the baseline (non-distillation) and previous SOTAs, $i.e.$, LabelEnc and LGD. The aforementioned two backbone, R-50 and R-101, and three detectors, FRCN, RetinaNet and FCOS, are selected to verify the performance of proposed mehtod on MS-COCO. Compared to the baseline model, our approach achieves about 3.5\% mAP gain for RetinaNet, 3.4\% mAP gain for FCOS and achieves notable advantages in other settings as well. Our method surpasses LabelEnc 2.8\% for FRCN-R50, 2.2\% for FRCN-R101 and more than 1\% over other settings. Our method can also achieve around 1\% gain for FRCN-R50 and surpasses LGD for FRCN and FCOS settings. To give more performance detail over the LGD, we involve more metrics for the comparison shown in Table~\ref{tab:lgd_table}. The performance of our proposed method in FRCN and FCOS outperform LGD in more AP related metrics. In RetinaNet$_{R101}$ setting, our proposed method achieve 5\% performance gain (26.1 vs 24.9) in APs. It is remarkable to mention that different from traditional knowledge distillation schemes, the teacher-free distillation method eliminates the uncertainty of finding a suitable teacher model.

\begin{table}[]
\caption{Comparisons with baseline and SOTA methods based on mean average precision (AP).}
\label{tab:sota_table}
\resizebox{1\columnwidth}{!}{
\begin{tabular}{clllll}
\hline
\multicolumn{1}{l}{Detector} & Backbone & Baseline & LabelEnc & LGD  & Ours \\ \hline
\multirow{2}{*}{FRCN}        & R-50     & 39.6     & 39.6     & 40.4 & \textbf{40.6} \\
                             & R-101    & 41.7     & 41.4     & 42.2 & \textbf{42.3} \\ \hline
\multirow{2}{*}{RetinaNet}   & R-50     & 38.8     & 39.6     & \textbf{40.3} & 40.2 \\
                             & R-101    & 40.6     & 41.5     & \textbf{42.1} & \textbf{42.1} \\ \hline
\multirow{2}{*}{FCOS}        & R-50     & 41.0     & 41.8     & 42.3 & \textbf{42.4} \\
                             & R-101    & 42.9     & 43.6     & 44.0 & \textbf{44.2 }\\ \hline
\end{tabular}}
\end{table}

\begin{table}[htbp]
\caption{Detailed Comparisons with LGD.}
\label{tab:lgd_table}
\resizebox{1\columnwidth}{!}{
\begin{tabular}{lrrrrrr}
\hline
 & \multicolumn{1}{l}{AP} & \multicolumn{1}{l}{AP50} & \multicolumn{1}{l}{AP75} & \multicolumn{1}{l}{APs} & \multicolumn{1}{l}{APm} & \multicolumn{1}{l}{APl} \\ \hline
FRCN$_{R50}$-Ours       & \textbf{40.6} & 61.2 & \textbf{44.0} & 23.8          & \textbf{43.9} & \textbf{53.2} \\
FRCN$_{R50}$-LGD        & 40.4          & \textbf{61.3} & 43.9          & \textbf{24.0}          & \textbf{43.9}          & 52.2          \\ \hline
FRCN$_{R101}$-Ours      & \textbf{42.3} & \textbf{62.9} & \textbf{45.8} & 25.3          & \textbf{45.9} & \textbf{56.3} \\
FRCN$_{R101}$-LGD       & 42.2          & 62.8          & 45.5          & \textbf{25.9}          & 45.5          & 56.0          \\ \hline
RetinaNet$_{R50}$-Ours  & 40.2          & 60.0          & \textbf{43.0} & \textbf{24.2} & \textbf{44.2} & 52.1          \\
RetinaNet$_{R50}$-LGD   & \textbf{40.3}          & \textbf{60.1}          & \textbf{43.0} & 24.0          & 44.1 & \textbf{52.4}         \\ \hline
RetinaNet$_{R101}$-Ours & \textbf{42.1} & 61.9          & 44.9          & \textbf{26.1} & 46.2          & \textbf{55.1} \\
RetinaNet$_{R101}$-LGD  & \textbf{42.1} & \textbf{62.1}          & \textbf{45.1}          & 24.9 & \textbf{46.5}         & 55.0 \\ \hline
FCOS$_{R50}$-Ours       & \textbf{42.4} & \textbf{61.2} & \textbf{46.0} & \textbf{26.4} & \textbf{46.1} & 54.0          \\
FCOS$_{R50}$-LGD        & \textbf{42.4} & \textbf{61.2} & 45.8          & 26.2          & \textbf{46.1}          & \textbf{54.3}          \\ \hline
FCOS$_{R101}$-Ours      & \textbf{44.2} & \textbf{63.3} & \textbf{47.6} & 27.1          & \textbf{48.3} & \textbf{57.5} \\
FCOS$_{R101}$-LGD       & 44.0          & 62.9          & 47.5          & \textbf{27.2}         & 47.9          & 56.9          \\ \hline
\end{tabular}
}
\end{table}

\subsection{Effect of Adjusting $\lambda$}
% TODO: say the point is to make sure lambda is comparable first, the we compare lambdas
Next, we considered the effect of varying the distillation coefficient, $\lambda$. 
While previous work assumed a $\lambda=1$~\cite{Zhang2022LGDLS}, we conjectured that adjusting $\lambda$ may be beneficial for model training due to varying the contribution of the distillation loss to the overall loss function during learning rate decay.
%Although the previous self-distillation approach, LGD, does not study the selection of the distillation loss coefficient, $\lambda$, we believe that a proper $\lambda$ is beneficial for model training. 
%The $\lambda$ determines the effect of knowledge distillation in training. 
Since we are using a different distillation loss than LGD, we first calibrated the $\lambda$ parameter between LGD and \SSSD{}.
First, we reproduced the original experiments by setting $\lambda=1$ in LGD with the FRCN detector and R50 backbone; the mean contribution of the penalized distillation loss to the total loss ($\lambda L_{distill}/L_{total}$) was 45\% after $1,000$ iterations.
%reproduce the experiments in LGD (fix $\lambda$ as 1) based on the FRCN-R50 setting and observe the value of $\lambda L_{distill}/L_{total}$ as a reference which is around 45\% in the early iterations. 
We computed a $\lambda$ in the domain of $[1,100]$ using binary search that yielded a mean $\lambda L_{distill}/L_{total} \approx 0.45$ after $1,000$ iterations, which led to an equivalent $\lambda$ of $50$ for \SSSD{}. 
To explore the impact of adjusting $\lambda$, we considered $\lambda \in \{1, 1.5, 2\}$ for LGD and $\lambda \in \{50, 75, 100\}$ for \SSSD{}. 
The $\lambda L_{distill}/L_{total}$ at iteration $17 \times 10^4$ was similar across the two architectures (Table~\ref{Tab:lambda}). %for the aforementioned experiments which helps us to understand the effect of knowledge distillation. 
Interestingly, the final $\lambda L_{distill}/L_{total}$ was close to $50\%$ for both LGD and \SSSD{} regardless of the $\lambda$. %One observation is that even the maximum $\lambda$ is 2$\times $ larger than the minimum $\lambda$ for both LGD and ours, the final $\lambda L_{distill}/L_{total}$ will be around 50\%. 

\begin{figure}[h] 
\centering
\includegraphics[width=1\columnwidth]{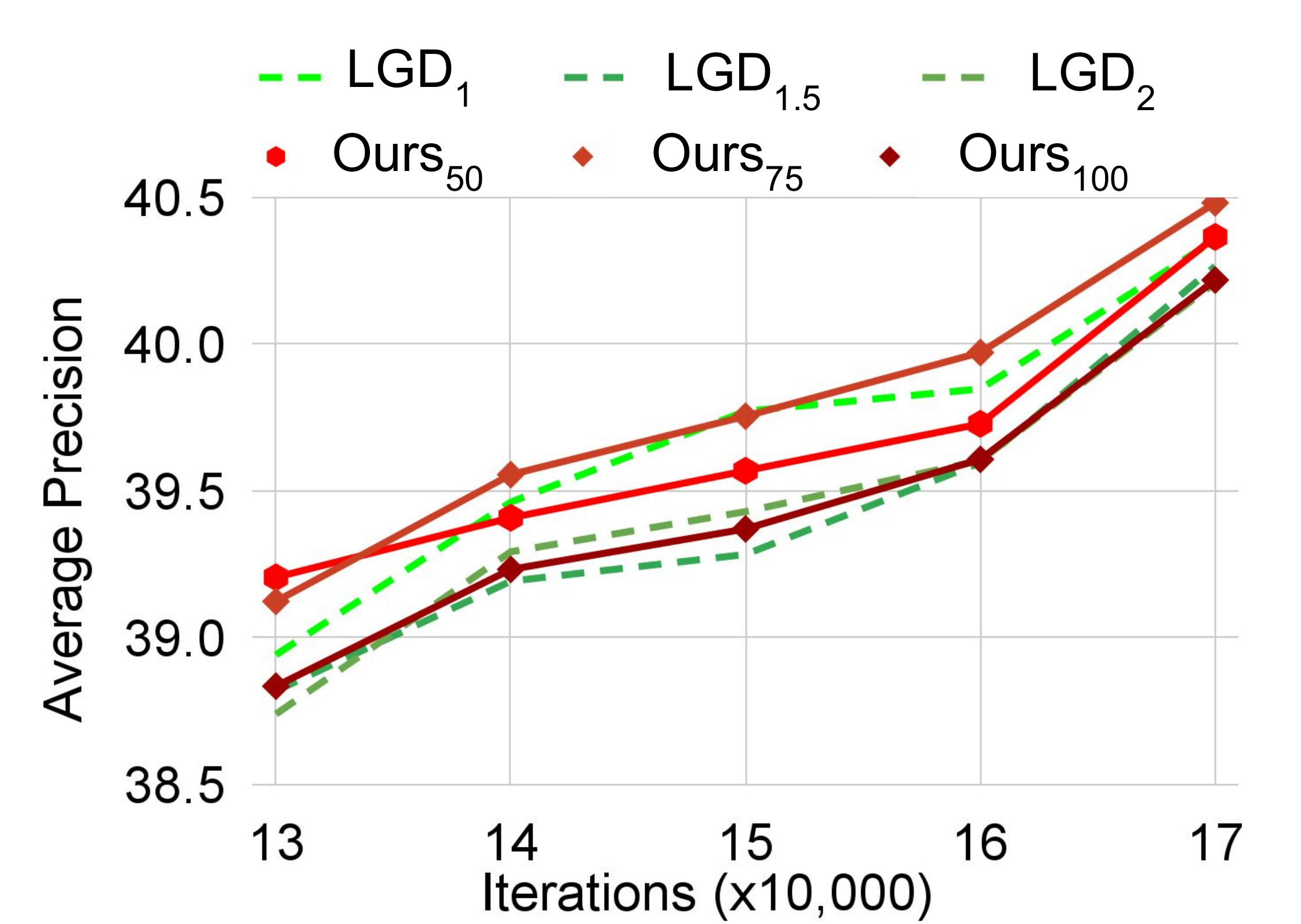}
\caption{\textbf{Performance comparison with different $\lambda$.} After calibrating the distillation loss, the AP for \SSSD{} with $\lambda=75$ (Ours$_{75}$) is higher than LGD configurations. The learning rates for each architecture are $0$ after iteration $17 \times 10^4$.}
\label{Fig:lambda}
\end{figure}

We compared the performance between LGD and \SSSD{} after calibrating $\lambda L_{distill}/L_{total}$ to be in a comparable range (Fig.~\ref{Fig:lambda} and Table~\ref{Tab:detailed_comparison_lambda}).
%According to Figure~\ref{Fig:lambda} and Table~\ref{Tab:detailed_comparison_lambda}, we compare the performance between LGD and ours when the $\lambda L_{distill}/L_{total}$ is in a comparable range as indicated in Table~\ref{Tab:lambda}. 
The top performing $\lambda$ for \SSSD{} ($75$) consistently outperformed the top performing LGD configuration ($\lambda=1$) in all AP measures besides AP50; when considering all $\lambda$, \SSSD{}  compares favorably to LGD among most of the AP variants, including up to a $1.1\%$ ($44.1$ versus $43.6$) improvement in AP75 (Table~\ref{Tab:detailed_comparison_lambda}). 
The top performing \SSSD{} also maintains an advantage over LGD from iterations $13 \times 10^4$ to $17 \times 10^4$ (Fig.~\ref{Fig:lambda}).
%Figure~\ref{Fig:lambda} also shows that Ours$_{75}$ is maintaining the lead over LGD$_{1}$ from $13 \times 10^4$ iterations to $17 \times 10^4$ iterations.

\begin{table}[]
\caption{Comparisons of $\lambda L_{distill}/L_{total}$ with different $\lambda$ after iterations $17 \times 10^4$.}
\begin{tabular}{llllll}
\hline
LGD$_1$                   & LGD$_{1.5}$                 & LGD$_2$                            & Ours$_{50}$                 & Ours$_{75}$                 & Ours$_{100}$                \\ \hline
\multicolumn{1}{c}{0.44} & \multicolumn{1}{c}{0.46} & \multicolumn{1}{c}{0.58} & \multicolumn{1}{c}{0.41} & \multicolumn{1}{c}{0.49} & \multicolumn{1}{c}{0.47} \\ \hline
\end{tabular}
\label{Tab:lambda}
\end{table}

\begin{table}[]
\caption{Detailed Comparisons with different $\lambda$ selections.}
\label{Tab:detailed_comparison_lambda}
\begin{tabular}{lrrrrrr}
\hline
     & \multicolumn{1}{l}{AP} & \multicolumn{1}{l}{AP50} & \multicolumn{1}{l}{AP75} & \multicolumn{1}{l}{APs} & \multicolumn{1}{l}{APm} & \multicolumn{1}{l}{APl} \\ \hline
LGD$_1$   & 40.4          & \textbf{61.3}            & 43.6            & 23.5                    & 43.7           & 53.1           \\
LGD$_{1.5}$  & 40.2                   & 60.8                     & 43.3                     & \textbf{23.8}           & 43.3                    & 52.7                    \\
LGD$_{2}$  & 40.2                   & 60.9                     & 43.6            & 23.5                    & 43.5                    & 53.0                    \\ \hline
Ours$_{50}$ & 40.5                   & 61.2            & \textbf{44.1}            & \textbf{23.8}           & 43.5                    & \textbf{53.2}           \\
Ours$_{75}$ & \textbf{40.6}          & 61.2            & \ 44.0            & \textbf{23.8}           & \textbf{43.9}           & \textbf{53.2}           \\
Ours$_{100}$ & 40.2                   & 60.7                     & 43.4                     & 23.5                    & 43.4                    & 52.6                    \\ \hline
\end{tabular}
\end{table}

\subsection{Stepwise Distillation}
\begin{figure}[h] 
\centering
\includegraphics[width=1\columnwidth]{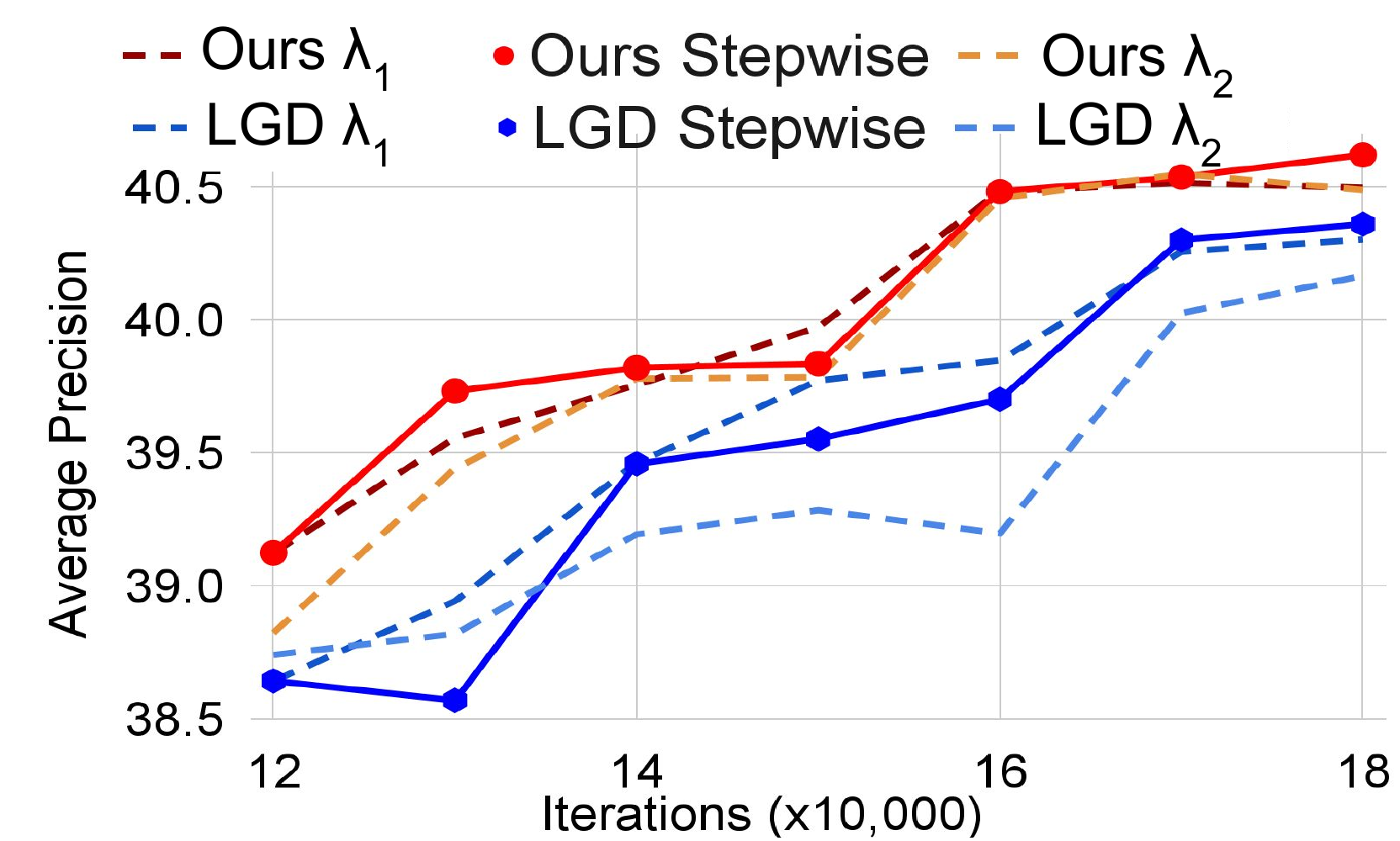}
\caption{\textbf{Stepwise self-distillation comparisons.}
The stepwise self-distillation strategy for both LGD and \SSSD{} (Ours) improves final AP over a fixed $\lambda$.}
\label{Fig:stepwise}
\end{figure}

Finally, we evaluated the effectiveness of stepwise distillation in both LGD and \SSSD{} using a fixed architecture (FRCN-R50) over the final $60,000$ iterations (Fig.~\ref{Fig:stepwise}).
We tested LGD $\lambda_1=1$ and \SSSD{} $\lambda_1=75$ since these were the best performing $\lambda$ for this architecture (Table~\ref{Tab:detailed_comparison_lambda}).
Additionally, we tested a slightly increased LGD $\lambda_2=1.5$ and \SSSD{} $\lambda_2=80$.
We compared these static $\lambda$ settings with stepwise distillation, which switches from $\lambda_1$ to $\lambda_2$ at iteration $120,000$ (in the learning rate scheduler period).
Stepwise distillation improves both LGD and \SSSD{} resulting in an approximately $0.5\%$ improvement in AP over fixed $\lambda$ settings (Fig.~\ref{Fig:stepwise}). 
%As shown in Figure~\ref{Fig:stepwise}, the curve shows that both groups, ours and LGD, are improved by stepwise distillation and the result shows that stepwise distillation reaches around 0.5\% performance gain over fixed $\lambda$ settings.
Since stepwise distillation does not impose additional computational costs and is independent of  the architecture, we optimistically believe that stepwise distillation may be beneficial for other knowledge distillation applications. 
%It is remarkable that the stepwise does not require extra cost in computing which is just simply adjusting the distillation coefficient $\lambda$ during learning rate decay. Since stepwise distillation is not dependent on any special designed structures or frameworks, we optimistically believe stepwise distillation may give a good inspiration for those studies on knowledge distillation.

\section{Conclusion}
In this paper, we proposed \textbf{S}mooth and \textbf{S}tepwise \textbf{S}elf-\textbf{D}istillation (\SSSD{}) for object detection, which can efficiently improve model performance without requiring a large teacher model.
Through extensive benchmarking, we demonstrated that \SSSD{} achieves improved performance when compared with current SOTA self-distillation approaches for a variety of backbones and detectors. %, including up to a 5\% performance gain over LGD. 
We investigated the effects of varying the distillation coefficient and justified stepwise distillation as a potentially beneficial procedure for improving the performance of knowledge distillation schemes. 
\label{sec:typestyle}
\bibliographystyle{IEEEbib}
\bibliography{refs}

\end{document}